\documentclass[11pt]{article}

\usepackage[margin=1in]{geometry}
\usepackage[T1]{fontenc}
\usepackage[utf8]{inputenc}
\usepackage{lmodern}
\usepackage{microtype}
\usepackage{amsmath,amssymb,amsthm,mathtools}
\usepackage{graphicx}
\usepackage{caption}
\usepackage{subcaption}
\usepackage{booktabs}
\usepackage{array}
\usepackage{tabularx}
\usepackage{multirow}
\usepackage{makecell}
\usepackage{float}
\usepackage{enumitem}
\usepackage{hyperref}
\usepackage{xcolor}
\usepackage{url}

\hypersetup{
  colorlinks=true,
  linkcolor=blue,
  citecolor=blue,
  urlcolor=blue,
  pdftitle={A Reproducibility Study of Partial Residual Ablations in Pre-LN Transformers},
  pdfauthor={Pratik Kumar Babariya}
}

\newcolumntype{Y}{>{\raggedright\arraybackslash}X}
\newcolumntype{P}[1]{>{\raggedright\arraybackslash}p{#1}}
\newcommand{\figref}[1]{Figure~\ref{#1}}

\title{A Reproducibility Study of Partial Residual Ablations in Pre-LN Transformers}
\author{Pratik Kumar Babariya\\Independent Research Project}
\date{}

\begin{document}
\maketitle

\begin{abstract}
I study what happens when attention-skip and FFN-skip connections are selectively removed from Pre-LN transformer blocks, training GPT-style models at 10M (TinyShakespeare) and 124M (OpenWebText) parameters. The central empirical result is an asymmetry. At 10M, a controlled 8-seed deterministic sweep gives AttnOnly $1.580 \pm 0.003$ and FFNOnly $3.350 \pm 0.002$. At 124M, a controlled five-seed matched sweep gives FullResidual $4.577 \pm 0.012$, AttnOnly $6.037 \pm 0.355$, and FFNOnly $7.569 \pm 0.001$; all five AttnOnly runs remain below all five FFNOnly runs, although AttnOnly remains substantially worse than FullResidual. I retain an intermediate 10M reproduction failure ($2.591 \pm 0.544$) and a runtime-gain confound as part of the reproducibility record. I then test a cross-position communication hypothesis at 10M using parameter-matched and windowed-attention interventions. A position-blind LocalMixer gives $2.490527 \pm 0.000369$, while self-only attention ($k=0$ previous tokens accessible) gives $2.490865 \pm 0.000849$. Allowing each attention layer to access one previous token in addition to the current token ($k=1$) changes the regime to $1.569693 \pm 0.006865$, a mean improvement of $0.921172$. Replicated radii $k=3,4,6,8,16,64$ remain in the lower-loss regime, with the best mean at $k=4$ ($1.517060 \pm 0.006029$). These interventions support a narrower mechanistic conclusion: in the tested 10M setting, cross-position communication is associated with the large recovery, but global context is not required; a per-layer radius of one preceding token produces most of the improvement. I report the experimental trail, including non-reproducing runs, for independent investigation.
\end{abstract}

\section{Introduction}
Residual connections, introduced by He et al.~\cite{he2016deep} for image classification, are now a defining structural feature of transformer language models~\cite{vaswani2017attention,radford2019language}. In the Pre-LN formulation each block applies two sublayers with identity shortcuts:
\begin{align}
 x^{k+1/2} &= x^k + \mathrm{Attn}(\mathrm{LN}_1(x^k)) \quad (\texttt{attn\_res = True}) \label{eq:attn} \\
 x^{k+1}   &= x^{k+1/2} + \mathrm{FFN}(\mathrm{LN}_2(x^{k+1/2})) \quad (\texttt{ffn\_res = True}) \label{eq:ffn}
\end{align}
The skip connections serve two purposes: they provide identity paths that can mitigate vanishing gradients in deep networks, and they allow each layer to learn residual transformations rather than full mappings~\cite{he2016deep}. The block Jacobian with a residual connection contains an identity term:
\begin{equation}
\frac{\partial x_{\ell+1}}{\partial x_\ell} = I + \frac{\partial f(x_\ell)}{\partial x_\ell}.
\end{equation}
This identity term provides a direct gradient path that can improve gradient propagation. Without it, gradients must traverse the sublayer Jacobian rather than receiving an additive identity contribution. Setting either residual flag to \texttt{False} removes the corresponding identity path, giving four configurations: Full Residual (both skips), AttnOnly (attention skip only), FFNOnly (FFN skip only), and No Residual (neither skip).

My initial experiments with a nanoGPT-style implementation reveal an apparently clean result: the two partial-residual configurations and No Residual reach nearly identical high-loss regimes, reaching validation loss $\approx$3.35 at 10M scale and $\approx$7.4 at 124M -- indistinguishable from No Residual. I treat this as a starting observation, not a conclusion, and ask whether the collapse is architectural or an artifact of the specific implementation and evaluation methodology.

\paragraph{Contributions.}
(1) A two-scale ablation study showing that the initial nanoGPT joint-collapse observation is implementation-sensitive rather than a sufficient final characterization. (2) Controlled ResidualGPT evidence for a robust residual-path asymmetry: at 10M, AttnOnly reaches $1.580 \pm 0.003$ across eight seeds while FFNOnly reaches $3.350 \pm 0.002$; at 124M, five matched seeds give FullResidual $4.577 \pm 0.012$, AttnOnly $6.037 \pm 0.355$, and FFNOnly $7.569 \pm 0.001$, with no overlap between the observed AttnOnly and FFNOnly ranges. (3) Identification and documentation of a runtime-gain/AdamW measurement confound and an intermediate cross-environment reproduction failure rather than omission of either result. (4) A controlled 10M mechanism study that varies cross-position communication while preserving the AttnOnly residual topology. Self-only attention gives $2.490865 \pm 0.000849$, whereas allowing one previous token per attention layer gives $1.569693 \pm 0.006865$ across the same three seeds. (5) A replicated communication-radius sweep showing that global context is not required for recovery: $k=3,4,6,8,16,64$ all remain in the lower-loss regime, with the best replicated mean at $k=4$. (6) Documentation of code, experimental results, and non-reproducing runs to support independent investigation.

\section{Background}
\subsection{Residual Networks and Gradient Flow}
He et al.~\cite{he2016deep} showed that identity shortcuts facilitate optimization of very deep networks by providing direct identity paths through residual blocks. The residual stream formalization by Elhage et al.~\cite{elhage2021framework} extended this to transformers: each sublayer reads from and writes to a shared vector that accumulates information across blocks. Removing a residual connection disrupts this compositional structure, not merely the optimization landscape.

\subsection{Initialization and Residual Coupling}
Standard initialization (std = 0.02) implicitly assumes the identity path dominates early training. T-Fixup~\cite{huang2020improving} and GPT-2 output scaling~\cite{radford2019language} demonstrate that initialization scale and residual connections are tightly coupled. When a residual is removed, this assumption breaks: the sublayer output is no longer stabilized by the identity path at step zero. This coupling motivated an initialization-sensitivity experiment during the reproducibility investigation; the first runtime-gain version of that experiment was later identified as confounded and is reported in Section 5.1.

\section{Experimental Setup}
\subsection{Two Implementations}
I report results from two implementations. The nanoGPT implementation (no weight tying, random validation batches at every eval, manual dot-product attention) was used for the initial joint-collapse observation at both scales. The ResidualGPT implementation uses weight tying between the token embedding and language model head, \texttt{F.scaled\_dot\_product\_attention} (flash attention), fixed validation batches (created once with seed 99{,}991 -- identical across all configurations), and per-layer activation storage for diagnostics. All asymmetry claims use ResidualGPT as the primary implementation. Implementation differences are explicitly reported wherever they affect results.

\subsection{Hyperparameters}
Table~\ref{tab:hyperparams} summarizes all hyperparameters.

\begin{table}[H]
\centering
\small
\caption{Hyperparameters for all experiments.}
\label{tab:hyperparams}
\resizebox{\textwidth}{!}{%
\begin{tabular}{|P{2.2cm}|P{2.1cm}|P{2.5cm}|P{6.3cm}|}
\hline
\textbf{Parameter} & \textbf{10M} & \textbf{124M} & \textbf{Shared} \\
\hline
n\_layer & 6 & 12 & lr: 3$\times$10$^{-4}$ (nanoGPT: constant; ResidualGPT 10M: cosine, 100-step warmup, floor 3$\times$10$^{-5}$) / 6$\times$10$^{-4}$ $\rightarrow$ 6$\times$10$^{-5}$ cosine, 2{,}000-step warmup (ResidualGPT 124M) \\
\hline
n\_embd & 384 & 768 & $\beta_1$=0.9, $\beta_2$=0.95, weight decay=0.1 \\
\hline
n\_head & 6 & 12 & Grad clip=1.0, dropout=0.2 (10M) / 0.1 (124M) \\
\hline
block\_size & 256 & 1024 & Optimizer: AdamW \\
\hline
Batch size & 64 & 16 & Grad accum: 1 (10M) / 8 (124M) \\
\hline
Train steps & 3{,}000 & 20{,}000 & Parameters $\approx$ 10.7M / 124.4M \\
\hline
Val interval & 100 & 500 & Dataset: TinyShakespeare / OpenWebText \\
\hline
\end{tabular}%
}
\end{table}

\subsection{Datasets}
The 10M experiments use TinyShakespeare with character-level tokenization (vocab=65, train=1.0M tokens, val=111K). The 124M experiments use OpenWebText~\cite{gokaslan2019openwebtext} with BPE tokenization via \texttt{tiktoken} (vocab=50,257, $\sim$113M tokens from 100,000 documents). Fixed validation batches (seed 99{,}991) are used in all ResidualGPT runs, ensuring identical evaluation data across all configurations.

\subsection{Configurations}
The residual-ablation sweeps compare four configurations: Full Residual (both skips active), No Residual (neither skip), AttnOnly (attention skip kept, FFN skip removed), and FFNOnly (FFN skip kept, attention skip removed). Within each sweep, training hyperparameters are held constant across configurations while residual flags vary. The separate mechanism experiments preserve the AttnOnly residual topology and vary the attention communication radius or replace attention with a position-wise LocalMixer.

\section{The Initial Observation: Joint Collapse}
Table~\ref{tab:nanogpt10m} shows the nanoGPT results at 10M scale. The two partial-residual configurations and No Residual reach nearly identical high-loss values to validation loss $\approx$3.35, a 2.27$\times$ degradation relative to Full Residual. This is the joint-collapse observation.

\begin{table}[H]
\centering
\small
\caption{nanoGPT implementation, 10M params, TinyShakespeare, seed 1337.}
\label{tab:nanogpt10m}
\begin{tabular}{|c|c|c|c|c|c|c|}
\hline
\textbf{Config} & \textbf{Attn} & \textbf{FFN} & \textbf{Best val} & \textbf{PPL} & \textbf{Final val} & \textbf{Ratio} \\
\hline
Full Residual & $\checkmark$ & $\checkmark$ & 1.4751 & 4.37 & 1.4751 & 1.00$\times$ \\
\hline
AttnOnly & $\checkmark$ & $\times$ & 3.3534 & 28.60 & 3.3534 & 2.27$\times$ \\
\hline
FFNOnly & $\times$ & $\checkmark$ & 3.3475 & 28.43 & 3.3475 & 2.27$\times$ \\
\hline
No Residual & $\times$ & $\times$ & 3.3523 & 28.57 & 3.3523 & 2.27$\times$ \\
\hline
\end{tabular}
\end{table}

\begin{figure}[H]
\centering
\includegraphics[width=\textwidth,trim=0bp 0bp 0bp 95bp,clip]{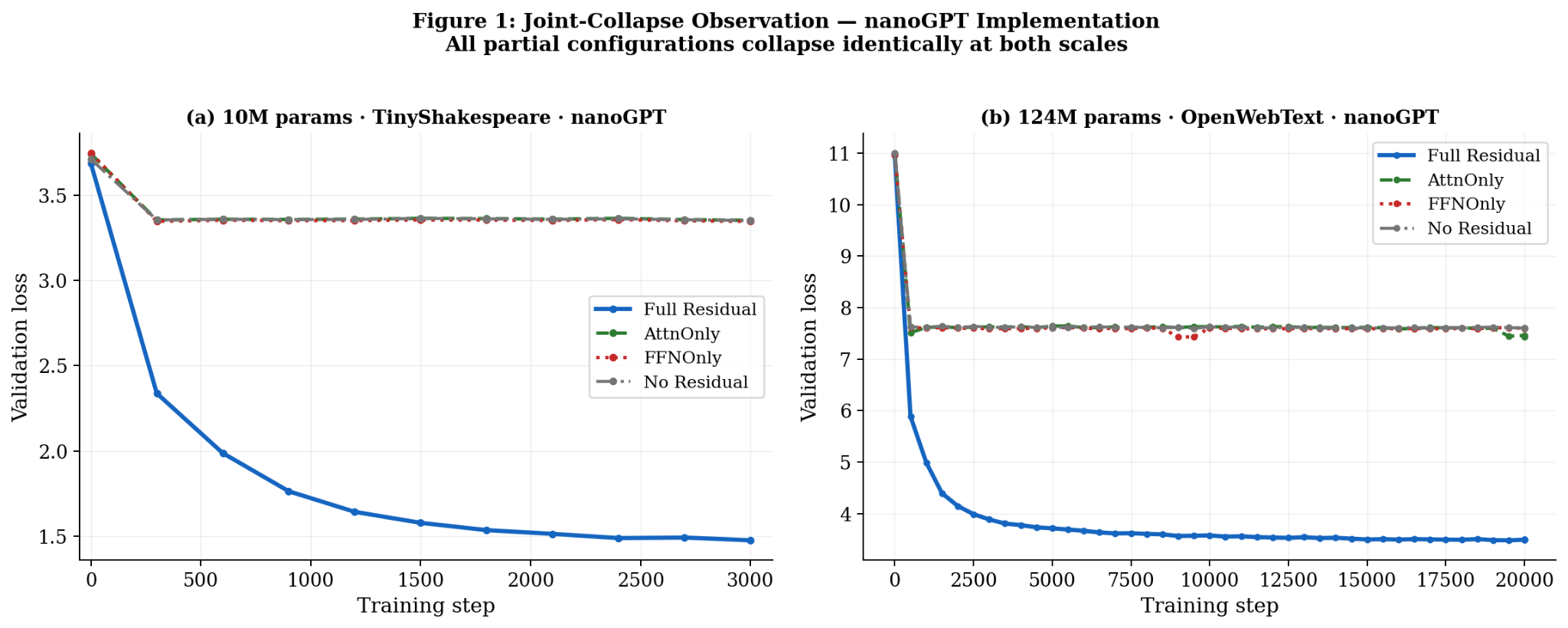}
\caption{Validation loss curves at 10M (left, TinyShakespeare) and 124M (right, OpenWebText) using the nanoGPT implementation. Both partial-residual configurations and NoResidual plateau in nearly the same high-loss regime at both scales.}
\label{fig:nanogptcurves}
\end{figure}

\figref{fig:nanogptcurves} shows val loss curves at both 10M and 124M nanoGPT scale. The pattern is identical: Full Residual descends to low loss while both partial-residual configurations and NoResidual plateau immediately after an initial drop. At 124M, degradation ratios (2.13$\times$ to 2.18$\times$) are consistent with 10M (2.27$\times$), confirming the joint-collapse observation is robust across scale, dataset, and tokenization scheme. In Table~\ref{tab:nanogpt10m}, Best val and Final val coincide for all four 10M configurations: validation loss was still monotonically improving at the last logged checkpoint (step 2,999) for each configuration, so the final measurement is also the best one observed during training.

\subsection{Mechanistic Signature of the Joint Collapse}
\begin{figure}[H]
\centering
\includegraphics[width=\textwidth,trim=0bp 0bp 0bp 65bp,clip]{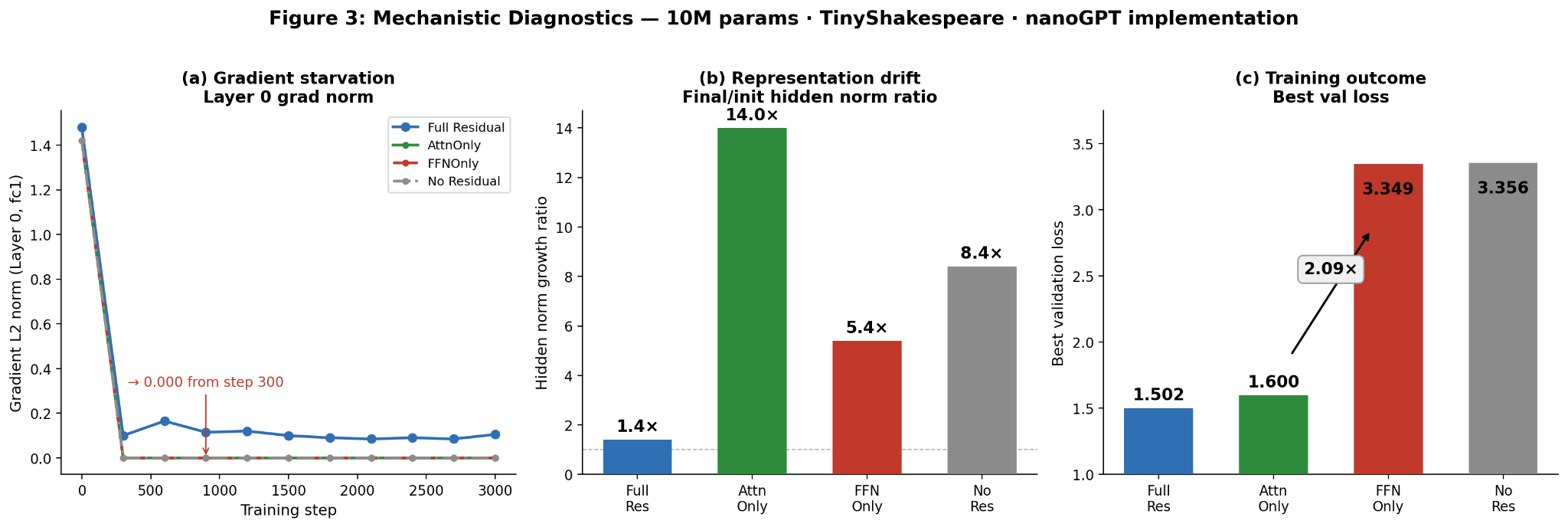}
\caption{Mechanistic diagnostics from the original 10M investigation. (a) In nanoGPT, the recorded Layer-0 fc1 parameter gradient norm rounds to 0.000 from step 300 onward for the ablated configurations while FullResidual remains nonzero. (b) Historical nanoGPT hidden-norm growth ratios: FullResidual 1.4$\times$, AttnOnly 14.0$\times$, FFNOnly 5.4$\times$, and NoResidual 8.4$\times$. (c) The originally recorded ResidualGPT outcomes: FullResidual 1.502, AttnOnly 1.600, FFNOnly 3.349, and NoResidual 3.356. Panel (c) does not show the later Phase-2 reproduction.}
\label{fig:mechanistic}
\end{figure}

Despite similar validation losses, the configurations show different internal behavior (\figref{fig:mechanistic}). The recorded Layer-0 gradient norm rounds to 0.000 from step 300 onward for all three ablated configurations, while Full Residual remains nonzero. Residual connections provide an additive identity contribution to the sublayer Jacobian, which can facilitate gradient propagation; this does not guarantee nonvanishing gradients through the entire model. Hidden-state norm growth ratios also differ: Full Residual grows 1.38$\times$, AttnOnly 14.03$\times$, and FFNOnly 5.37$\times$. These differences motivate further investigation but do not alone identify the causes of the observed failures.

\section{Challenging the Observation}
\subsection{A Measurement Confound: Runtime Scaling vs. Weight Initialization}
My first attempt to test initialization sensitivity applied gain as a runtime multiplier: output = g $\times$ sublayer(x) at every forward pass, for g $\in$ \{0.25, 0.5, 2.0, 4.0\}. At g = 2.0, AttnOnly appeared to recover to best val 1.54. I identified this as a confound before reporting the result: AdamW's per-parameter adaptive learning rate compensates for a constant multiplicative reparameterization within $\approx$200 steps, making all gain values functionally equivalent. Loss curves across all gain values overlapped completely (total spread $<$ 0.002 at any step). The apparent recovery was an optimizer artifact, not a genuine initialization effect. I corrected the implementation to use weight-level reinitialization applied before optimizer construction, and rebuilt the experiment. This confound is reported here in full as a methodological contribution: it is easy to miss, produces convincing-looking results, and requires careful experimental design to avoid.

\subsection{Implementation Difference: Fixed Validation Batches}
The nanoGPT implementation evaluates on randomly sampled validation batches at every checkpoint. For a slowly-learning model such as AttnOnly, this introduces high evaluation variance: a model that is genuinely improving can appear to have plateaued under noisy evaluation. ResidualGPT creates fixed validation batches once at startup (seed 99{,}991) and reuses the identical samples at every evaluation across all four configurations. This eliminates evaluation-sampling noise as one implementation difference between the two systems, but it does not isolate the cause of the nanoGPT-to-ResidualGPT change. I verify that \texttt{set\_global\_seed()} does not reseed the batcher generator, confirming fixed val batches are identical across all four configs. This explanation is not formally isolated from other differences between the two implementations: ResidualGPT also differs from nanoGPT in weight tying between the token embedding and output head, use of flash attention (\texttt{F.scaled\_dot\_product\_attention}) versus a manual implementation, and minor optimizer/initialization details. I have not run an ablation that holds all of these fixed except validation batching, so no single implementation difference is assigned causal priority; any of the bundled differences, individually or in combination, could be responsible.

\section{The Central Observation: Controlled Residual-Path Asymmetry at 10M and 124M}
\subsection{10M Scale --- ResidualGPT}
Table~\ref{tab:resgpt10m} shows the ResidualGPT primary results at 10M scale, as originally recorded: AttnOnly was measured below the collapse floor while FFNOnly collapsed. A later intermediate reproduction on different hardware did not recover this statistic, but a subsequent controlled 8-seed deterministic sweep confirmed the 10M effect: AttnOnly mean 1.580 $\pm$ 0.003 across 8 seeds, FFNOnly mean 3.350 $\pm$ 0.002, gap of 1.770 val loss units, no seed approaching the floor. See Section~\ref{sec:reprohistory} for the full three-phase account.

\begin{table}[H]
\centering
\small
\caption{ResidualGPT implementation, 10M params, TinyShakespeare. Historical mean $\pm$ population standard deviation across three seeds for AttnOnly and FFNOnly; FullResidual and NoResidual are single-seed anchors. AttnOnly values are as originally recorded; see Section~\ref{sec:reprohistory} for a subsequent reproduction attempt that did not recover this exact statistic.}
\label{tab:resgpt10m}
\begin{tabular}{|P{2.7cm}|P{2.6cm}|P{1.2cm}|P{2.0cm}|P{2.7cm}|}
\hline
\textbf{Config} & \textbf{Best val loss $\pm$ std} & \textbf{PPL} & \textbf{Seeds} & \textbf{vs. Full Residual} \\
\hline
Full Residual & 1.5017 -- & 4.49 & 1337 & -- (baseline) \\
\hline
AttnOnly & $1.5999 \pm 0.0149$ & 4.95 & 1337, 42, 123 & +6.5\% [ORIGINALLY RECORDED BELOW FLOOR] \\
\hline
FFNOnly & $3.3493 \pm 0.0009$ & 28.48 & 1337, 42, 123 & +123.0\% [COLLAPSES] \\
\hline
No Residual & 3.3564 -- & 28.69 & 1337 & +123.5\% [COLLAPSES] \\
\hline
\end{tabular}
\end{table}

\begin{figure}[H]
\centering
\includegraphics[width=0.96\textwidth]{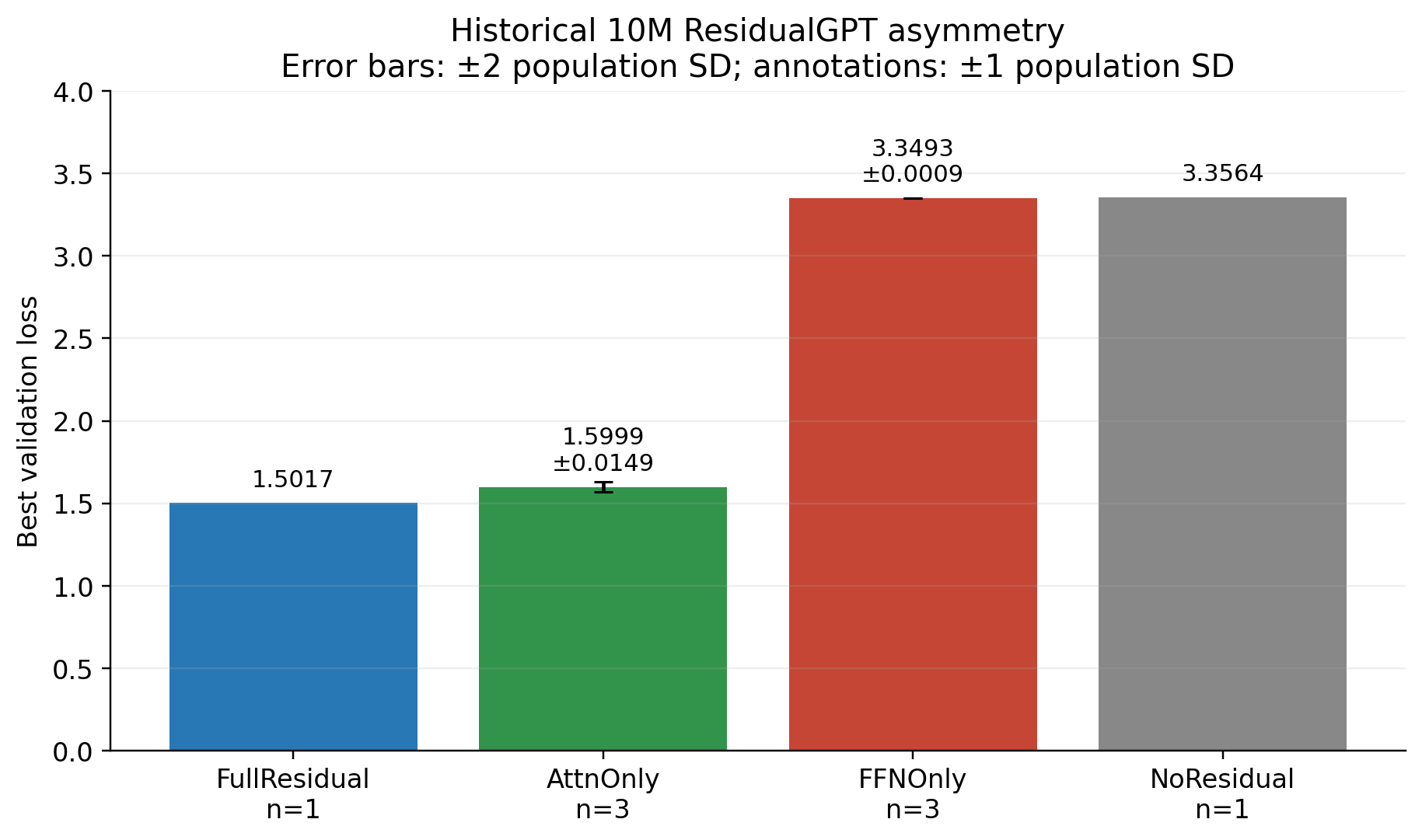}
\caption{Historical 10M ResidualGPT asymmetry as originally recorded. The bars show FullResidual 1.5017, AttnOnly $1.5999 \pm 0.0149$ across three seeds, FFNOnly $3.3493 \pm 0.0009$ across three seeds, and the NoResidual seed-1337 anchor 3.3564. This figure does not visualize the later reproduction attempt; that reproduction is retained in the text as experimental history. Numerical annotations report mean $\pm$ population SD, whereas plotted error bars extend to $\pm 2$ population SD for the multi-seed configurations. FullResidual and NoResidual are single-seed anchors.}
\label{fig:resgpt10m}
\end{figure}

\figref{fig:resgpt10m} visualizes the originally recorded 10M asymmetry. A subsequent reproduction did not recover the original AttnOnly statistic and included one seed near the collapse floor (see Section~\ref{sec:reprohistory}); that reproduction is retained in text rather than plotted here. The historical FFNOnly/AttnOnly loss ratio is 2.09$\times$. The historical FFNOnly mean and NoResidual anchor differ by approximately 0.007 loss units. They are numerically close, but the single NoResidual run does not establish uncertainty for this difference. The historical population standard deviations are 0.0009 for FFNOnly and 0.0149 for AttnOnly, indicating lower observed seed variation for FFNOnly in this three-seed comparison. These statistics do not characterize the separate Phase-2 reproduction, whose reported AttnOnly summary was $2.591 \pm 0.544$.

\begin{figure}[H]
\centering
\includegraphics[width=0.96\textwidth]{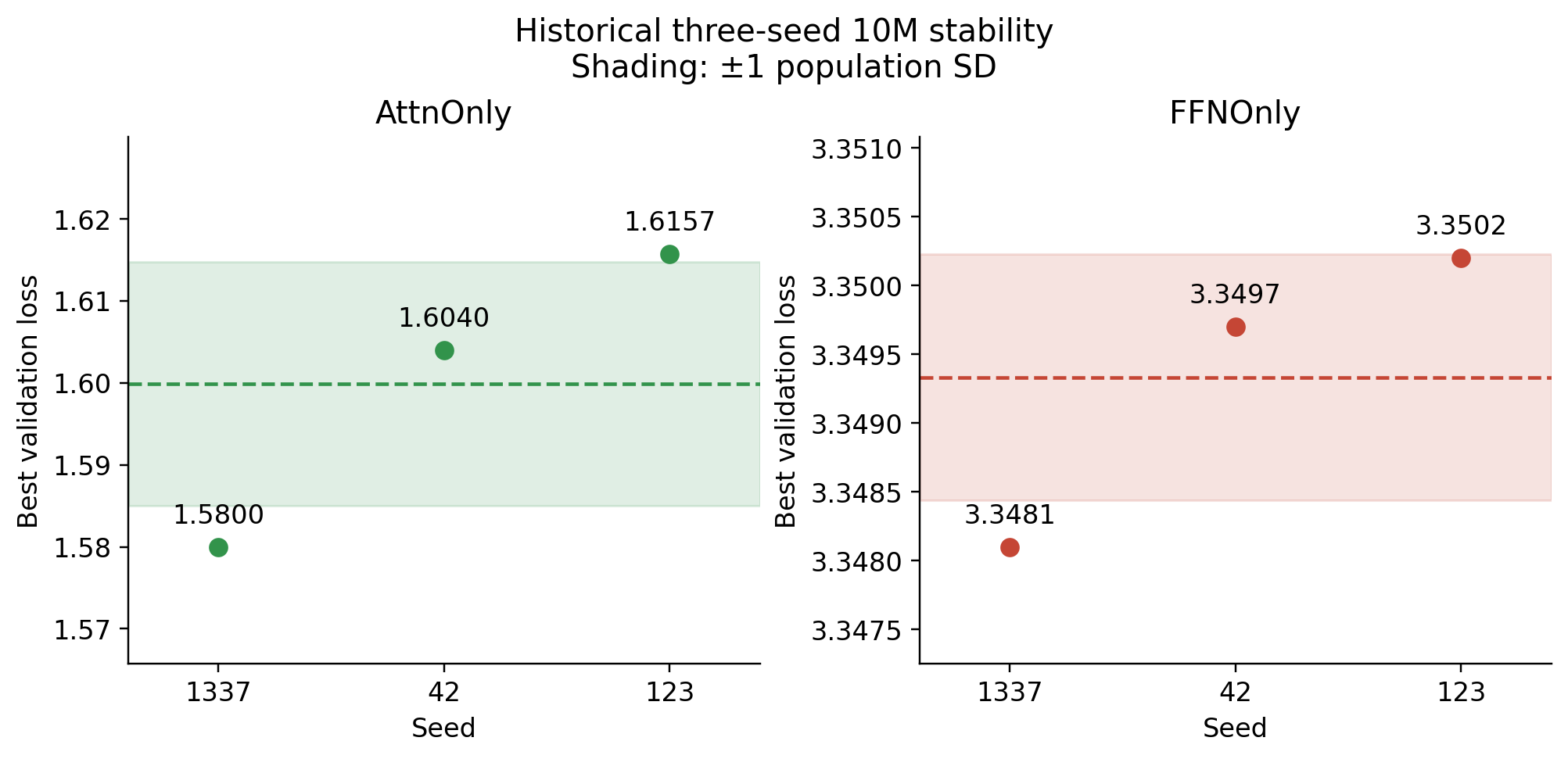}
\caption{Historical three-seed 10M ResidualGPT stability result as originally recorded. AttnOnly gives mean 1.5999 $\pm$ 0.0149 across seeds 1337, 42, and 123, while FFNOnly gives 3.3493 $\pm$ 0.0009. Individual points show each seed, and dashed lines show the means. This figure is retained as experimental history; the later controlled Phase-3 sweep in Figure~\ref{fig:phase3} provides the primary 10M estimate. Historical standard deviations use the population convention; shaded bands represent $\pm 1$ population SD.}
\label{fig:historical3seed}
\end{figure}

\begin{figure}[H]
\centering
\includegraphics[width=0.90\textwidth]{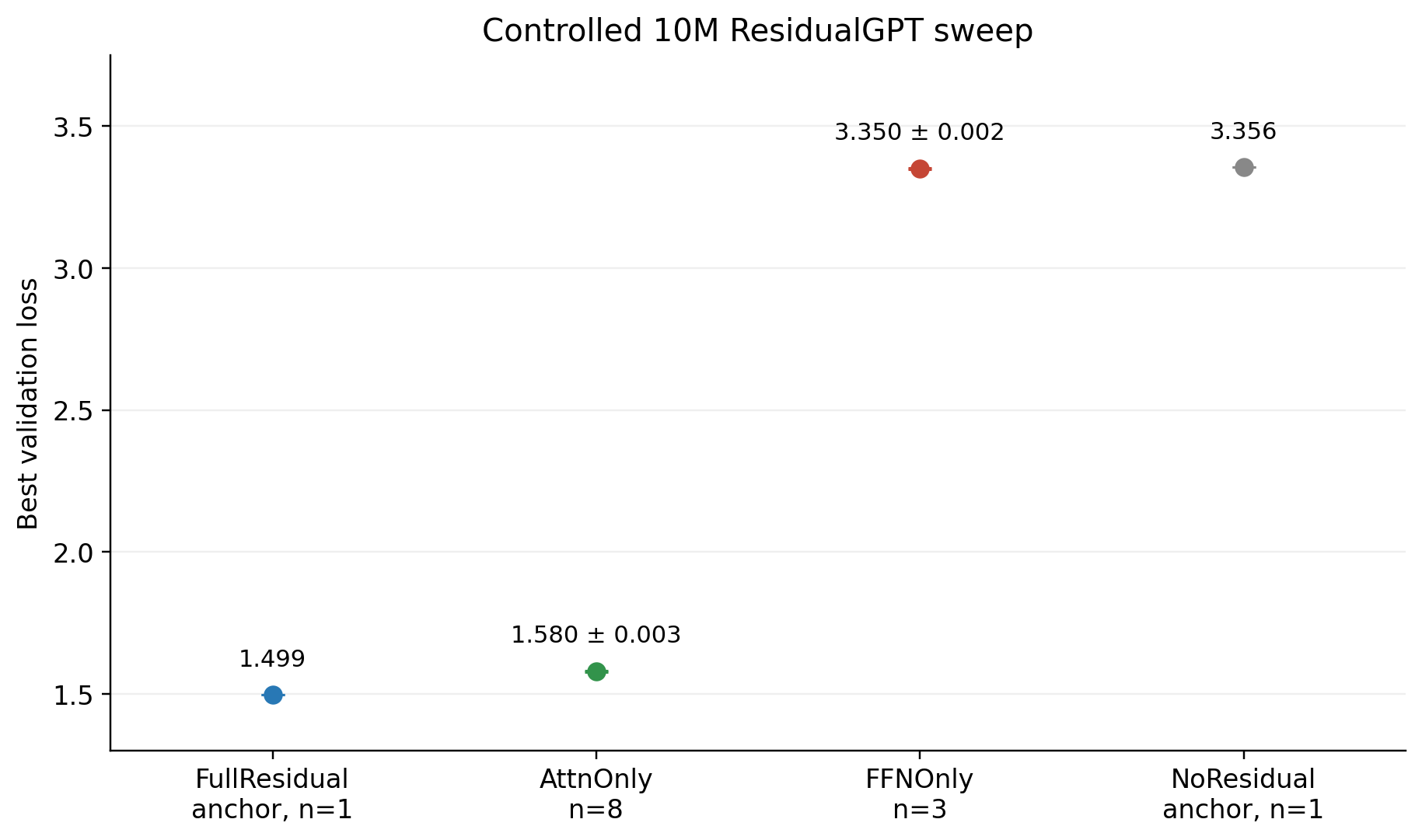}
\caption{Controlled Phase-3 10M ResidualGPT confirmation. Points show the reported aggregate best-validation-loss statistics: FullResidual anchor 1.499, AttnOnly 1.580 $\pm$ 0.003 across eight seeds, FFNOnly 3.350 $\pm$ 0.002 across three seeds, and NoResidual anchor 3.356. Error bars denote sample standard deviation where multi-seed statistics are available. The AttnOnly--FFNOnly mean separation is 1.770 validation-loss units. Individual eight-seed values are not reconstructed here because the retained paper artifacts support the aggregate statistics rather than all per-seed values.}
\label{fig:phase3}
\end{figure}

\subsection{124M Scale Verification --- ResidualGPT}
The original 124M seed-1337 run is retained as historical evidence in Table~\ref{tab:resgpt124m}. Because the historical and controlled 124M runs were produced under different documented environments, their absolute losses should not be treated as a single sweep. The primary 124M evidence is therefore the within-environment ranking from the later matched-seed study.

\begin{table}[H]
\centering
\small
\caption{Historical ResidualGPT 124M results on OpenWebText, seed 1337. These values are retained unchanged as experimental history and are not pooled with the later controlled sweep.}
\label{tab:resgpt124m}
\begin{tabular}{|P{2.8cm}|P{2.1cm}|P{1.2cm}|P{3.1cm}|}
\hline
\textbf{Config} & \textbf{Best val loss} & \textbf{PPL} & \textbf{Ratio / Result} \\
\hline
Full Residual & 3.2630 & 26.13 & 1.00$\times$ / Baseline \\
\hline
AttnOnly & 4.8994 & 134.21 & 1.50$\times$ / originally below floor (+50.1\% above Full Res) \\
\hline
FFNOnly & 7.6662 & 2,135 & 2.35$\times$ / high-loss regime \\
\hline
No Residual & 7.6887 & 2,184 & 2.36$\times$ / floor \\
\hline
\end{tabular}
\end{table}

\begin{figure}[H]
\centering
\includegraphics[width=0.96\textwidth]{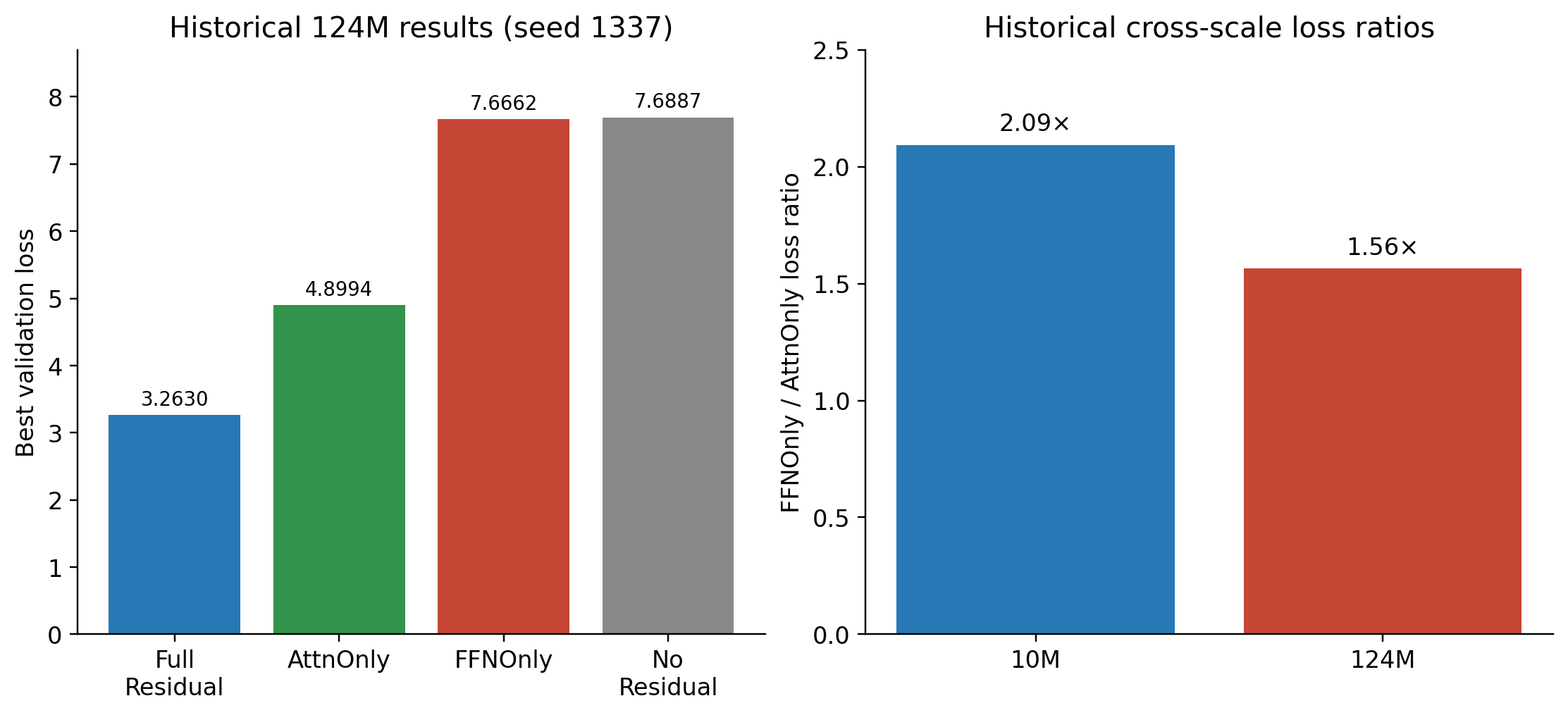}
\caption{Historical 124M ResidualGPT scale-verification figure retained as experimental history. Panel (a) shows the originally recorded seed-1337 comparison (FullResidual 3.2630, AttnOnly 4.8994, FFNOnly 7.6662, NoResidual 7.6887); panel (b) shows the historical FFNOnly/AttnOnly loss ratios (2.09 at 10M and 1.56 at 124M). These values are not pooled with the later controlled matched-seed sweep.}
\label{fig:scale124_historical}
\end{figure}

\subsubsection{Controlled 124M matched-seed sweep}
To resolve the high variance in the historical AttnOnly runs, I ran a matched-seed sweep using seeds 1337, 42, 123, 0, and 7 for FullResidual, AttnOnly, and FFNOnly in one controlled environment (A100-SXM4-40GB, PyTorch 2.11.0+cu128, CUDA 12.8). The training budget remained 20,000 steps and the comparison metric remained best validation loss. A contemporaneous NoResidual seed-1337 anchor was interrupted at step 18,000, after its best validation loss had occurred at step 2,000.

\begin{table}[H]
\centering
\small
\caption{Controlled 124M ResidualGPT sweep on OpenWebText. Mean $\pm$ sample standard deviation across five matched seeds (1337, 42, 123, 0, 7). NoResidual is a contemporaneous seed-1337 anchor.}
\label{tab:resgpt124m_controlled}
\begin{tabular}{|P{2.5cm}|P{2.4cm}|c|P{2.4cm}|P{4.2cm}|}
\hline
\textbf{Config} & \textbf{Best val loss} & \textbf{$n$} & \textbf{Range} & \textbf{Interpretation} \\
\hline
Full Residual & $4.577 \pm 0.012$ & 5 & 4.560--4.592 & Baseline; tight across seeds \\
\hline
AttnOnly & $6.037 \pm 0.355$ & 5 & 5.662--6.508 & Degraded vs. baseline, but below FFNOnly in every matched run \\
\hline
FFNOnly & $7.569 \pm 0.001$ & 5 & 7.568--7.570 & Tight high-loss regime \\
\hline
No Residual & 7.584 & 1 & -- & Seed-1337 floor anchor \\
\hline
\end{tabular}
\end{table}

\begin{figure}[H]
\centering
\includegraphics[width=0.90\textwidth]{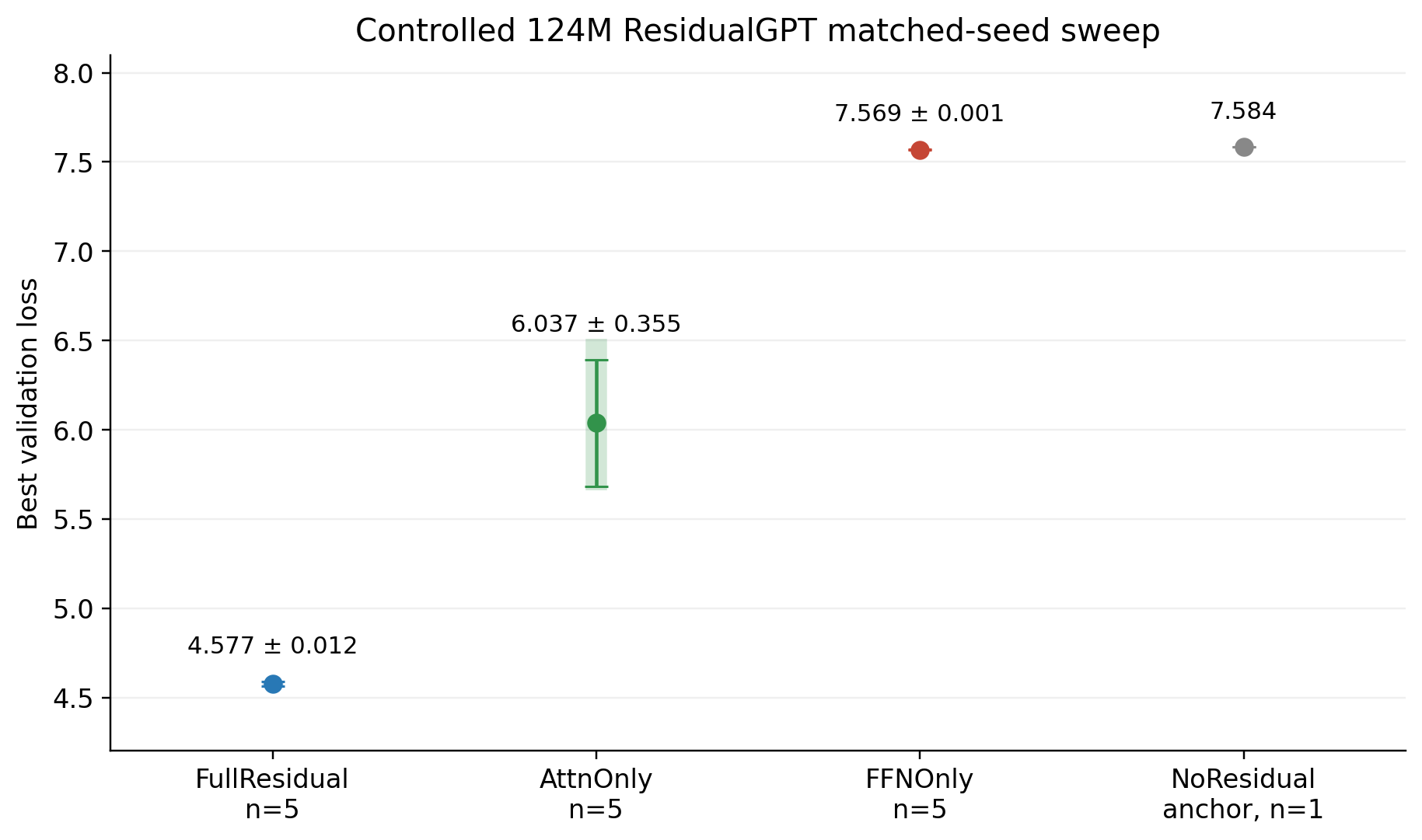}
\caption{Controlled 124M ResidualGPT matched-seed summary. Points show mean best validation loss; error bars show sample standard deviation, and the lighter vertical bars show the observed min--max range for the five-seed FullResidual, AttnOnly, and FFNOnly sweeps. NoResidual is the contemporaneous seed-1337 anchor. The figure uses reported aggregate and range statistics and does not reconstruct unavailable individual seed values. The closest observed AttnOnly--FFNOnly pair is separated by 1.060 validation-loss units.}
\label{fig:controlled124}
\end{figure}

The mean AttnOnly--FFNOnly separation is 1.532 validation-loss units. The observed ranges do not overlap: the worst AttnOnly seed reaches 6.508 while the best FFNOnly seed reaches 7.568, leaving a minimum observed gap of 1.060. FFNOnly is also close to the contemporaneous NoResidual anchor (7.569 versus 7.584, difference 0.015). Because NoResidual has one seed in this sweep, it is treated as an anchor rather than a distribution. These results establish that the direction of the asymmetry persists at 124M under matched seeds, while AttnOnly remains substantially worse than FullResidual and is much more seed-sensitive than at 10M.

\subsection{Cross-Scale Summary}
Table~\ref{tab:crossscale} separates historical observations from the controlled sweeps. Absolute losses differ across documented environments, so within-sweep rankings are the primary evidence.

\begin{table}[H]
\centering
\small
\caption{Cross-scale asymmetry summary. Historical rows are retained unchanged; controlled rows provide the primary multi-seed evidence. The historical 10M ResidualGPT row retains population SDs; controlled-sweep SDs follow the conventions stated in their respective tables and captions.}
\label{tab:crossscale}
\resizebox{\textwidth}{!}{%
\begin{tabular}{|c|l|c|c|c|c|}
\hline
\textbf{Scale} & \textbf{Impl.} & \textbf{AttnOnly best val} & \textbf{FFNOnly best val} & \textbf{AttnOnly gap} & \textbf{Asym. ratio} \\
\hline
10M & nanoGPT & 3.353 (high-loss) & 3.348 (high-loss) & +127.3\% & 1.00$\times$ \\
\hline
10M & ResidualGPT historical & $1.600\pm0.015$ & $3.349\pm0.001$ & +6.5\% & 2.09$\times$ \\
\hline
10M & ResidualGPT Phase 3 (8-seed sweep) & $1.580\pm0.003$ & $3.350\pm0.002$ & +5.4\% & 2.12$\times$ \\
\hline
124M & nanoGPT & 7.433 (high-loss) & 7.430 (high-loss) & +113.5\% & 1.00$\times$ \\
\hline
124M & ResidualGPT historical & 4.899 & 7.666 & +50.1\% & 1.56$\times$ \\
\hline
124M & ResidualGPT controlled sweep & $6.037\pm0.355$ & $7.569\pm0.001$ & +31.9\% & 1.25$\times$ \\
\hline
\end{tabular}%
}
\end{table}

The direction of the asymmetry is supported by controlled multi-seed sweeps at both scales. At 10M, AttnOnly reaches $1.580 \pm 0.003$ while FFNOnly reaches $3.350 \pm 0.002$. At 124M, AttnOnly reaches $6.037 \pm 0.355$ while FFNOnly reaches $7.569 \pm 0.001$, and all five AttnOnly seeds remain below all five FFNOnly seeds. The scale change is not a uniform loss shift: FullResidual and FFNOnly remain tight across the five 124M seeds, whereas AttnOnly becomes substantially more seed-sensitive. This establishes the direction of the asymmetry at 124M but does not by itself explain the increased AttnOnly variability.

\section{Mechanistic Intervention: Communication Radius}
The empirical asymmetry is better established than its mechanism. I therefore test a specific structural hypothesis: whether cross-position communication in the surviving attention pathway is associated with AttnOnly recovery. These experiments use the clean 10M deterministic environment and preserve the AttnOnly residual topology (attention skip on, FFN skip off).

\subsection{Interventions and Definition of \texorpdfstring{$k$}{k}}
The communication-radius experiments restrict causal attention so that position $t$ may attend only to positions
\begin{equation}
 t-k \le j \le t,
\end{equation}
Here $k$ is the number of previous tokens accessible within each attention layer, in addition to the current token. The radius is defined per layer, not for the entire model. Information can propagate across multiple positions through stacked layers; therefore, $k=1$ does not restrict the model's total receptive field to one preceding token. Thus $k=0$ is strictly self-only attention, $k=1$ permits $\{t-1,t\}$, and larger $k$ values expand the local causal window. A pre-run single-attention-module visibility test for $k=1$ used \texttt{WindowedSelfAttention(window\_size=1)} in evaluation mode. Perturbing position 0 affected query position 1 (maximum absolute change 6.189413) but produced no measured effect at positions 2 and 3 (maximum absolute change 0.0). This test verifies the module-level attention mask, not the full stacked model's receptive field. All windowed models retain 10,750,080 parameters and use the same 3,000-step schedule, fixed validation batches, optimizer, initialization, and residual flags.

As a complementary no-communication control, I also replace attention with a parameter-matched position-wise LocalMixer. The LocalMixer has the same parameter count as the attention module but cannot exchange information across token positions.

\subsection{No Communication vs. One Previous Token}
The central result is the boundary between self-only computation and the smallest nonzero communication radius. Across matched seeds 1337, 42, and 123, self-only attention gives
\begin{equation}
\boxed{k=0: 2.490865 \pm 0.000849},
\end{equation}
while allowing one previous token per attention layer gives
\begin{equation}
\boxed{k=1: 1.569693 \pm 0.006865}.
\end{equation}
The mean improvement is 0.921172 validation-loss units. The paired improvements are 0.928037, 0.919061, and 0.916419 for seeds 1337, 42, and 123 respectively. The independently parameter-matched LocalMixer gives $2.490527 \pm 0.000369$, nearly identical to the self-only attention mean (difference 0.000338). Thus two distinct interventions that remove cross-position communication occupy the same degraded regime, whereas introducing access to one preceding token per attention layer produces a large recovery.

\subsection{Replicated Communication-Radius Sweep}
Table~\ref{tab:radius} summarizes the replicated radius sweep. The values are means $\pm$ sample standard deviations across seeds 1337, 42, and 123. The $k=2$ result is retained separately as a single-seed fine-grained measurement and is not included in the replicated summary.

\begin{table}[H]
\centering
\small
\caption{10M communication-radius mechanism study. Here $k$ is the number of previous tokens accessible per attention layer in addition to the current token. All reported rows use three matched seeds. Individual seed values are rounded for display; summary statistics retain the notebook-reported values computed from the underlying run results. Lower validation loss is better.}
\label{tab:radius}
\begin{tabular}{|c|P{3.2cm}|c|c|c|c|}
\hline
\textbf{$k$} & \textbf{Accessible context} & \textbf{Seed 1337} & \textbf{Seed 42} & \textbf{Seed 123} & \textbf{Mean $\pm$ std} \\
\hline
0 & self only & 2.490095 & 2.490724 & 2.491776 & $2.490865 \pm 0.000849$ \\
\hline
1 & self + 1 previous & 1.562058 & 1.571663 & 1.575357 & $1.569693 \pm 0.006865$ \\
\hline
3 & self + 3 previous & 1.527295 & 1.527308 & 1.517451 & $1.524018 \pm 0.005687$ \\
\hline
4 & self + 4 previous & 1.518006 & 1.522560 & 1.510614 & $1.517060 \pm 0.006029$ \\
\hline
6 & self + 6 previous & 1.520973 & 1.525632 & 1.520637 & $1.522414 \pm 0.002792$ \\
\hline
8 & self + 8 previous & 1.521617 & 1.539797 & 1.526760 & $1.529392 \pm 0.009371$ \\
\hline
16 & self + 16 previous & 1.537717 & 1.524254 & 1.534447 & $1.532139 \pm 0.007022$ \\
\hline
64 & self + 64 previous & 1.561206 & 1.555237 & 1.571378 & $1.562607 \pm 0.008161$ \\
\hline
\end{tabular}
\end{table}

\begin{figure}[H]
\centering
\includegraphics[width=0.90\textwidth]{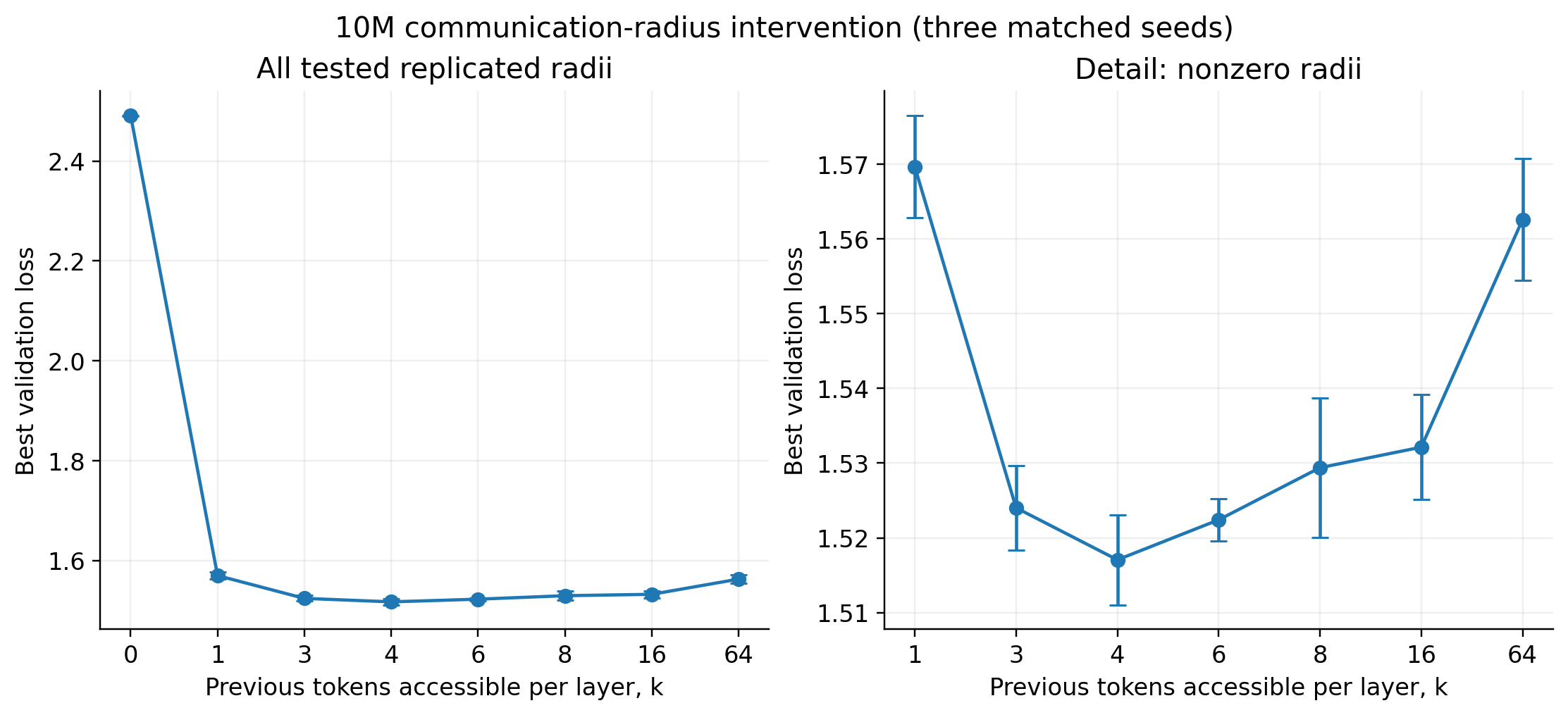}
\caption{Replicated 10M communication-radius intervention. Points show mean best validation loss across seeds 1337, 42, and 123; error bars show sample standard deviation. The left panel shows all replicated radii; the right panel enlarges the nonzero-radius results. Radii are displayed at equally spaced categorical positions. The dominant transition occurs from $k=0$ (self only) to $k=1$ (one previous token accessible per attention layer). Increasing the radius beyond a small local window gives comparatively modest changes and does not improve monotonically.}
\label{fig:radius}
\end{figure}

A single-seed $k=2$ run gives 1.535775 at seed 1337, consistent with the sharp early improvement but not promoted to a replicated estimate. The replicated curve reaches its lowest tested mean at $k=4$ ($1.517060 \pm 0.006029$); across the subsequently tested replicated radii $k=6,8,16,64$, the mean increases to $1.562607 \pm 0.008161$ by $k=64$. I treat the non-monotonic radius dependence descriptively: the present experiments establish that global context is not required for recovery, but they do not identify why a small local window performs best under this 3,000-step training budget.

\subsection{Mechanistic Interpretation}
These interventions strengthen the cross-position communication account while narrowing it. The result is not that AttnOnly requires global causal attention. Rather, within the tested 10M setup, eliminating cross-position communication produces a degraded regime near 2.49 in both self-only attention and a position-wise LocalMixer, while permitting access to one preceding token per attention layer produces most of the observed recovery. Larger local windows yield smaller additional changes. This is causal evidence about the tested intervention family, not a universal theorem about transformers or a complete explanation of the residual asymmetry at other scales.

\section{Discussion}
\subsection{Claims and Reproducibility}\label{sec:reprohistory}
The primary 10M asymmetry estimate remains the Phase-3 controlled result: AttnOnly $1.580 \pm 0.003$ across eight seeds versus FFNOnly $3.350 \pm 0.002$ across three controls, separated by 1.770 validation-loss units. Earlier 10M values remain experimental history rather than being pooled with this sweep. At 124M, the historical AttnOnly values 4.899, 7.586, and 5.967 motivated the matched sweep; the controlled five-seed result is FullResidual $4.577 \pm 0.012$, AttnOnly $6.037 \pm 0.355$, and FFNOnly $7.569 \pm 0.001$, with a minimum observed AttnOnly--FFNOnly gap of 1.060.
For completeness, the earlier 10M reproduction history is retained separately. Phase-1 repeated runs at seed 1337 produced losses 1.97, 1.94, and 2.06. Phase-2 runs at seeds 1337, 42, and 123 produced approximately 2.289, 3.354, and 2.130; the previously reported summary was $2.591 \pm 0.544$. These values are retained as historical reports rather than recomputed estimates; the historical summary's SD convention should not be assumed to match the sample SD convention used in the new intervention tables. The position-blind mixer tested in the Phase-2 environment reached 2.487. That experiment is distinct from the new clean-environment LocalMixer study, which gives $2.490527 \pm 0.000369$. The cause of the cross-environment discrepancy remains unresolved.

The new mechanism study is a separate 10M intervention series. Its strongest matched result is the $k=0$ to $k=1$ boundary: $2.490865 \pm 0.000849$ versus $1.569693 \pm 0.006865$. The parameter-matched LocalMixer independently gives $2.490527 \pm 0.000369$. These results support the claim that cross-position communication is associated with the recovery in this tested setting, while the replicated radius sweep shows that global context is not required.

\subsection{Research Practice}
The runtime-gain confound and the failed intermediate reproduction are retained because they materially changed the experimental design. A multiplicative forward-pass gain produced convincing apparent recovery before I identified that AdamW absorbed the reparameterization within approximately 200 steps. I discarded that interpretation, switched to weight-level initialization, preserved the failed reproduction attempts, and used controlled sweeps and matched interventions to separate robust findings from environment-specific observations.

\subsection{Limitations}\label{sec:limitations}
The claims are bounded in several ways. Training budgets are 3K steps at 10M and 20K at 124M, so the results characterize these budgets rather than asymptotic convergence. The nanoGPT-to-ResidualGPT transition bundles fixed validation batches, weight tying, attention implementation, and minor optimizer/initialization changes; their individual effects are not isolated. The study covers the tested Pre-LN GPT-style models only; Post-LN, RMSNorm, SwiGLU, and RoPE variants are untested. The controlled 124M NoResidual evidence is one seed-1337 anchor, interrupted at step 18,000, not a distribution. The 10M Phase-3 sweep and mechanism experiments use A100-PCIE-40GB, PyTorch 2.12.0+cu130, CUDA 13.0, while the controlled 124M sweep uses A100-SXM4-40GB, PyTorch 2.11.0+cu128, CUDA 12.8; accordingly, within-sweep comparisons are primary and absolute losses are not equated across environments. No formal hypothesis test is used, and the available seed counts do not characterize distribution tails. Finally, the communication-radius mechanism is established only at 10M on TinyShakespeare; whether the same sharp $k=0$ to $k=1$ transition appears at 124M or in modernized transformer blocks remains open.
The communication interventions establish effects within the AttnOnly configuration. They do not by themselves establish that these effects are specific to residual removal or fully explain the AttnOnly--FFNOnly asymmetry. Matched communication-radius sweeps under other residual configurations would help distinguish a general benefit of contextual information from an effect specific to the ablated residual topology.

He \& Hofmann~\cite{hehofmann2024} independently report that naive attention-skip removal harms trainability and that MLP-skip removal causes substantial slowdown without the same catastrophic behavior. Their work asks whether architectural modifications can make skip removal harmless, whereas this study measures unmodified Pre-LN blocks and then intervenes on communication radius to probe the observed asymmetry.

\section{Conclusion}
Selective residual removal is asymmetric in the controlled ResidualGPT experiments. At 10M, the primary Phase-3 sweep gives AttnOnly $1.580 \pm 0.003$ and FFNOnly $3.350 \pm 0.002$, separated by 1.770 validation-loss units. At 124M, five matched seeds give FullResidual $4.577 \pm 0.012$, AttnOnly $6.037 \pm 0.355$, and FFNOnly $7.569 \pm 0.001$; every observed AttnOnly run remains below every FFNOnly run, with a minimum gap of 1.060, while the single NoResidual anchor reaches 7.584. The result is therefore narrower than ``AttnOnly is good'': removing the FFN skip remains substantially worse than FullResidual but avoids the high-loss regime reached by attention-skip removal at both tested scales.

The 10M mechanism experiments sharpen this empirical result. Self-only attention ($k=0$) gives $2.490865 \pm 0.000849$, nearly identical to a parameter-matched position-wise LocalMixer at $2.490527 \pm 0.000369$. Allowing each attention layer to access one previous token in addition to the current token ($k=1$) gives $1.569693 \pm 0.006865$, a mean improvement of 0.921172. Replicated radii from $k=3$ through $k=64$ remain in the lower-loss regime, with the best mean at $k=4$. Thus, in the tested 10M setting, the evidence supports cross-position communication as a central ingredient in the recovery while showing that global context is unnecessary: a per-layer radius of one preceding token produces most of the transition. The remaining questions concern why $k=4$ has the lowest observed mean among the tested replicated radii under this training budget, whether the same boundary persists at 124M, and how broadly the result generalizes beyond the tested Pre-LN architecture.

\section*{Code Availability}
The project repository is available on \href{https://github.com/pratik376/Why-Partial-Residuals-fails}{GitHub (pratik376/Why-Partial-Residuals-fails)}. This statement identifies the repository and does not assert that every experiment history or checkpoint is included in the public release.

\end{document}